%% file: main.tex
\documentclass[letterpaper]{article} % DO NOT CHANGE THIS
\usepackage{aaai2026}  % DO NOT CHANGE THIS
\nocopyright
\usepackage{times}  % DO NOT CHANGE THIS
\usepackage{helvet}  % DO NOT CHANGE THIS
\usepackage{courier}  % DO NOT CHANGE THIS
\usepackage[hyphens]{url}  % DO NOT CHANGE THIS
\usepackage{graphicx} % DO NOT CHANGE THIS
\usepackage{natbib}  % DO NOT CHANGE THIS AND DO NOT ADD ANY OPTIONS TO IT
\usepackage{caption} % DO NOT CHANGE THIS AND DO NOT ADD ANY OPTIONS TO IT
\usepackage{algorithm}
\usepackage{algorithmic}
\usepackage{newfloat}
\usepackage{listings}
\DeclareCaptionStyle{ruled}{labelfont=normalfont,labelsep=colon,strut=off} % DO NOT CHANGE THIS
\floatstyle{ruled}
\newfloat{listing}{tb}{lst}{}
\floatname{listing}{Listing}
\title{Generative AI Alignment with Hinduism's Theological Plurality and Sacred Representation}
\author{
    Dipto Das\textsuperscript{\rm 1}, Arpita Kundu\textsuperscript{\rm 2}, Nusrat Jahan Mim\textsuperscript{\rm 3}, Shion Guha\textsuperscript{\rm 4}, Syed Ishtiaque Ahmed\textsuperscript{\rm 1}
}
\affiliations{
    \textsuperscript{\rm 1} Department of Computer Science, \textsuperscript{\rm 3} Faculty of Architecture, Landscape and Design, \textsuperscript{\rm 4} Faculty of Information,\\  University of Toronto\\
    \textsuperscript{\rm 2} Khulna University of Engineering \& Technology\\
    dipto.das@utoronto.ca, arpitakundu309@gmail.com, nusrat.mim@daniels.utoronto.ca, shion.guha@utoronto.ca, ishtiaque@cs.toronto.edu
}

\usepackage[T1]{fontenc}
\usepackage{xspace}
\usepackage[para]{footmisc}

\input{words}
\begin{document}

\maketitle

\begin{abstract}
    Generative AI systems are increasingly used to answer personal questions and mediate everyday practices, including religion. However, existing discussions around AI alignment and ethics have largely centered secular, Western, and Abrahamic assumptions about religion, offering limited attention to other faith-based traditions. In this paper, we examine how Hindu users engage with generative AI systems in relation to their religious knowledge, belief, and practice. Drawing on 15 semi-structured interviews with Bangladeshi Hindu participants, we analyze how users interpret AI-generated religious representations, scriptural explanations, devotional interactions, and synthetic religious media. We found that AI can be both accessible and ethically troubling. While AI supported scriptural inquiry, devotional visualization, and religious storytelling, our study also identified concerns about theological flattening, cultural misrepresentation, devotional manipulation, and the simulation of sacred presence and authority. We conclude by arguing that religious alignment in generative AI requires interpretive alignment: systems that disclose their limits, preserve plurality, and avoid simulating sacred authority and sycophantic personalization.
\end{abstract}

\input{sections/introduction}%\clearpage
\input{sections/literature_review}%\clearpage
\input{sections/methods}%\clearpage
\input{sections/results}%\clearpage
\input{sections/discussion}%\clearpage
\input{sections/conclusion}%\clearpage
\input{sections/optional_statements}

\bibliography{aaai2026}

\end{document}

%% file: sections/introduction.tex
\section{Introduction}
Generative AI (GenAI) systems are increasingly becoming sites where people ask personal, moral, and existential questions~\cite{howe2023chatgpt, cole2025artificial, krugel2023chatgpt}. These uses now extend into religion, as evident in the emergence of religion-specific chatbots and AI scripture companions, including BibleGPT\footnote{\url{https://biblegpt-la.com/}}, QuranGPT\footnote{\url{https://www.qurangpt.com/}}, GitaGPT\footnote{\url{https://gitagpt.org/}}, DharmaChat\footnote{\url{https://dharmachat.in/}}, Faith Explorer\footnote{\url{https://faithexplorer.app/}}, Hadith AI\footnote{\url{https://www.hadith-ai.com/}}, BiblePics\footnote{\url{https://www.biblepics.co/}}, and Text With Jesus\footnote{\url{https://textwith.app/jesus/}}. But are they \emph{aligned} with respective religious values and the principles of religious co-existence~\cite{rifat2026homeroom}, and what does it mean for a GenAI system to be ``aligned" in a religious context? This question is difficult because religion is not only a domain of factual knowledge or group identity; it is also interpretive, embodied, affective, relational, and contested.

Prior work on AI ethics and alignment has largely approached harm through concerns such as factual inaccuracy~\cite{augenstein2024factuality}, offensive representation~\cite{gehman2020realtoxicityprompts}, stereotyping~\cite{caliskan2017semantics}, deception~\cite{park2024ai}, and overtrust~\cite{maynez2020faithfulness}. These concerns are important, but they are insufficient for understanding the use of AI in religious contexts. GenAI systems do not merely retrieve or display religious information. They also synthesize, visualize, personalize, and simulate religious presence~\cite{nooreyezdan2023india}. As a result, the ethical problem is not only whether an output is accurate or offensive, but also which interpretive tradition it privileges, which lived contexts it erases, and whether it presents generated content as if it carried sacred authority. Therefore, the existing approaches to alignment and representation risk missing how AI systems may appear religiously inclusive while still flattening religious plurality and reshaping devotional meaning.

Our starting insight is that religious AI use must be understood as an interpretive problem rather than only a representational one. In this study, we examined the case of Hinduism, the world's third-largest religion and the largest major non-Abrahamic religious tradition~\cite{vaughan2020most}. Hindu religious life is constituted through multiple interpretive traditions, regional practices, and situated forms of authority~\cite{leach2014religion}. Since GenAI systems tend to produce confident, singular, and personalized outputs~\cite{farquhar2024detecting, cheng2026sycophantic}, religious AI systems are likely to make theological and cultural choices even when they appear neutral. Therefore, alignment in this context must mean designing systems that distinguish between explanation and authority and avoid collapsing religious plurality into a single optimized answer. Particularly, we focused on Bangladeshi Hindus. While Hinduism is often represented through India-centric templates~\cite{sinha2006problematizing}, Bangladesh has one of the world's largest Hindu populations ($\sim$13.1 million)~\cite{bbs2022preliminary}, and Bengali Hindu practice there is shaped by regional devotional traditions, colonial histories, and minority life~\cite{rifat2024politics}. To understand Hindu users' experiences with AI-generated religious images, scriptural explanations, devotional storytelling, synthetic religious media, and chatbot-based spiritual guidance, we conducted 15 semi-structured interviews with Bangladeshi Hindu participants.

We analyzed how Hindu users evaluated AI systems in relation to religious representation, authenticity, authority, caste, regional specificity, and devotional practice. Beyond asking whether they found AI outputs useful or harmful, we examined how they interpreted those outputs through Hindu theological concepts, Bengali Hindu practices, caste and regional experience, and broader histories of religious representation. In doing so, we traced patterns and situated differences in how Bangladeshi Hindu users understood religious meaning, authority, and harm. We found that AI systems can support scriptural inquiry, devotional visualization, and religious storytelling, especially for users without extensive scriptural training. However, the same systems also produced aesthetic distortion, theological flattening, cultural misrecognition, devotional manipulation, and blasphemous simulations of divine presence. These findings lead to three broader reflections. First, GenAI can reproduce colonial and majoritarian representational logics through sanitized, exoticized, or monolithic templates. Second, GenAI's anthropomorphic tendency takes on a distinct ethical charge in religious contexts when systems simulate divine interactions. Third, religious alignment should be understood as interpretive rather than based on group identity. Overall, such systems should help users navigate plurality with context and humility, not speak as authoritative religious agents.

%% file: sections/literature_review.tex
\section{Literature Review}
We review three bodies of work that together frame our study: research on GenAI as moral and spiritual companions, AI alignment and harm in religious contexts, and the challenges posed by Hinduism's internally plural traditions.

\subsection{Generative AI as Moral and Spiritual Companions}
AI has an anthropomorphic tendency~\cite{shneiderman2023ai}. Studies of social and companion chatbots, such as Replika, show that users can develop attachments to conversational agents through anthropomorphism, perceived authenticity, social presence, and repeated interaction~\cite{pentina2023exploring}. In moral domains, ChatGPT's advice has been shown to influence users' judgments even when they know it comes from a chatbot~\cite{krugel2023chatgpt}. These findings are especially important in religious contexts, where users may approach AI systems not only for factual explanations but also for guidance on suffering, duty, devotion, sin, ritual, or moral uncertainty. While conversational agents may support reflection, encouragement, and spiritual growth, but should not occupy the role of clergy, pastors, priests, or other religious authorities~\cite{cole2025artificial, trothen2022replika}. Related work on religion-based AI chatbots similarly notes that these systems answer theological and practical questions by drawing on scriptures such as the Bible, Quran, Bhagavad Gita, and Torah, but may also produce inaccurate answers, homogenized religious accounts, and gendered biases~\cite{biana2024feminist}. Tsuria and Tsuria further show that GenAI tools can present moralized accounts of religion, raising questions about how AI systems translate religious traditions into normative and comparative frameworks~\cite{tsuria2024artificial}. Hence, we should examine how AI systems are increasingly designed and encountered as spiritual companions, moral advisors, and interpretive mediators.

Beyond text-based chatbots, AI is also reshaping religious life through synthetic sacred media. Work on religious robots has examined machines that perform blessings or represent divine and clerical figures, showing how technologies can make religion newly visible through ritualized and anthropomorphic forms~\cite{midson2022posthuman, balle2020robots}. While text-to-image (T2I) systems can visualize sacred narratives, these biblical images also raise questions about theological accuracy, aesthetic convention, cultural bias, and the limits of machine-generated imagination~\cite{van2025mimesis, makimei2025seeing}. Beyond assessing whether a generated image or answer is accurate, the more complex ethical stakes lie in how synthetic sacred media reshape the conditions through which religious figures, scriptures, and practices become perceptible, memorable, emotionally compelling, and authoritative.

Compared to Christianity and Islam, scholars have only begun to address how these systems operate across non-Abrahamic, internally plural, and regionally situated religious traditions, such as Hinduism. Much of the emerging discussion focuses on whether AI can support spirituality, improve access to scripture, or avoid misinformation and bias~\cite{biana2024feminist, tsuria2024artificial, zhang2025cognitive}. Less attention has been paid to how GenAI systems choose among competing interpretive traditions, how they simulate sacred presence, or how they transform devotional relationships and spiritual authority through synthetic voice, image, and personalization. We extend this literature by examining how Bangladeshi Hindus interpret AI-generated religious images, scriptural explanations, devotional videos, and chatbot-based spiritual guidance.

\subsection{AI Alignment, Harm, and Religious Contexts}
AI alignment research asks how AI systems can be designed to act in accordance with human intentions, preferences, interests, or values. However, alignment is not only a technical problem but also a normative one: different accounts of alignment imply different assumptions about whose values matter, how disagreement should be handled, and what kind of human judgment AI systems should defer to~\cite{gabriel2020artificial}. This is especially important for GenAI systems, where harms often arise not only from system failure but from plausible, fluent, and socially persuasive outputs~\cite{bender2021dangers, ji2023survey}. Therefore, prior work on language model risk has examined harms such as discrimination, stereotypes, exclusion, toxicity, misinformation, inconsistency, manipulation, over-reliance, and other human-computer interaction risks~\cite{weidinger2022taxonomy, gehman2020realtoxicityprompts, krugel2023chatgpt}.

These harm categories are crucial, but religious contexts expose their limits. When users ask religious questions, they are often asking for more than just factual information. They may be seeking moral orientation, devotional meaning, scriptural interpretation, ritual guidance, or reassurance during uncertainty~\cite{pargament2000many, cummings2010medicine}. Recent work on sycophancy and anthropomorphism further complicates alignment in religious contexts. Large language models (LLMs) optimized through human feedback may learn to agree with or affirm users' stated views rather than challenge false, harmful, or exclusionary assumptions~\cite{sharma2024towards, cheng2026sycophantic}. At the same time, prior work on anthropomorphic AI shows that users may attribute mind, agency, social presence, or trustworthiness to computational systems, especially when systems communicate through human-like conversational cues~\cite{waytz2014mind, pentina2023exploring, mcguire2023reputational}. In ordinary advice settings, sycophancy and anthropomorphism can encourage dependence, overtrust, or poor judgment. In religious settings, the stakes can be different: agreeable, human-like responses may validate orthodoxy or personal bias under the guise of personalization, while also making the system appear to be a spiritual interlocutor rather than a limited computational artifact~\cite{loewen2022fabulation, cole2025artificial}. Thus, religious alignment should not be reduced to matching a user's preferences or representing a community's group identity. A system that simply adapts to what a user asks for may become more persuasive precisely when it should disclose uncertainty, refuse harmful framings, or point to interpretive disagreement.

Considering these perspectives together, we treat religious AI harm as an interpretive problem as much as a representational, informational, or interactional one. Existing AI ethics frameworks help identify problems of toxicity, stereotyping, misinformation, deception, overtrust, sycophancy, and anthropomorphic attachment, but they are less equipped to explain harms that arise when AI systems simulate sacred authority, collapse theological plurality, or present one tradition's interpretation as religion-as-such. This paper builds on these discussions by asking what alignment means when the relevant domain is not a stable set of facts or preferences, but a contested field of scriptures, practices, embodied traditions, regional histories, and devotional relationships.

\subsection{Hinduism, Plurality, and Representation}
What is now called ``Hinduism" is better understood as a plural and internally differentiated set of traditions than as a single unified doctrine~\cite{nicholson2013unifying}. Hindu traditions comprise diverse philosophical systems, textual traditions, ritual practices, devotional lineages, and regional interpretations~\cite{frazier2012ritual}. A useful entry point into this plurality is the distinction between \emph{shruti} (``that which is heard") and \emph{smriti} (``that which is remembered")~\cite{leach2014religion}. \emph{Shruti} texts, especially the \emph{Vedas} and \emph{Upanishads}, are often treated as foundational and of divine origin, while \emph{smriti} includes later interpretive, narrative, ritual, and social traditions. Among the most influential \emph{smriti} texts are the \emph{Mahabharata} and the \emph{Ramayana}. The \emph{Mahabharata} contains the \emph{Bhagavad Gita} (commonly referred to as \emph{Gita} in short), a dialogue between Krishna and Arjuna on duty (\emph{dharma}), action, and devotion (\emph{bhakti}), while the \emph{Ramayana} narrates the life of Rama and ideals of moral order. Alongside these epics, the \emph{Puranas} elaborate cosmologies, genealogies, and deity-centered devotional traditions. These texts continue to shape everyday Hindu religious imagination, ritual practice, storytelling, temple iconography, and devotional imagery~\cite{lutgendorf1991life, leach2014religion}. Concepts such as \emph{karma}, rebirth, temporary heavens and hells, and liberation (\emph{moksha}) recur across these traditions, but their meanings and relative importance vary across interpretive contexts.

Hindu plurality is also philosophical, devotional, and regional~\cite{flood1996introduction}. Major Vedantic schools such as \emph{Advaita} or non-dualism, \emph{Vishishtadvaita} or qualified non-dualism, and \emph{Dvaita} or dualism offer different understandings of the relationship among self, ultimate reality, and the world~\cite{betty2010dvaita}. These philosophical traditions coexist with devotional formations centered on deities such as Vishnu, Shiva, Devi, Ganesha, and Surya, often understood through both \emph{sakara} (with form) and \emph{nirakara} (formless) conceptions of divinity~\cite{aktor2017hindu}. Hindu religious life is also mediated through localized devotional lineages and \emph{bhakti} traditions centered around gurus, saints, and preachers, such as Chaitanya in Bengal, Kabir in North India, and Sai Baba in Western India~\cite{sen1998bhakti, hess2015bodies, white1972sai}. These traditions illustrate why Hindu representation cannot be reduced to scriptural reference or visual recognizability alone. A depiction of a deity, an explanation of caste, or an interpretation of the \emph{Gita} always draws from some situated theological, linguistic, regional, or devotional frame.

These issues of representation are historically mediated. Colonial and postcolonial processes significantly reshaped how Hinduism came to be represented and understood~\cite{sugirtharajah2004imagining, bhatia2017unforgetting}. Orientalist scholarship, missionary critique, and print circulation encouraged models of religion centered on singular authoritative texts, coherent doctrines, and recognizable belief systems~\cite{pennington2005hinduism}. Practicing Hindus and intellectuals also reinterpreted traditions within these conditions, contributing to new forms of religious self-understanding and nationalist identity formation~\cite{nicholson2013unifying}. These transformations continue into the present, as contemporary political and cultural projects, including Hindu nationalist movements, shape which texts, deities, practices, and narratives are foregrounded as representative of Hinduism~\cite{mehta2022hindu, nandy1989intimate}. Social categories like caste are similarly represented through generalized or various theological frameworks, even though caste is lived differently across regional and social contexts~\cite{gupta2022caste, dirks1989original}. Hence, we approach Hinduism not as a fixed object to be represented by AI, but as a historically mediated and internally plural formation shaped by layered interpretations, regional practices, social hierarchies, and ongoing political contestation.

%% file: sections/methods.tex
\section{Methods}
We conducted a qualitative study between June 2024 and April 2026 to understand how Bangladeshi Hindu users interpret and engage with GenAI systems in relation to religious knowledge, belief, and practice. Given the importance of situated and context-dependent interpretations in these communities, we sought to understand participants' experiences, reflections, and meaning-making processes.

% \subsection{Data Collection}
We recruited participants through a combination of convenience, purposive, and snowball sampling~\cite{goodman1961snowball, golzar2022convenience, suri2011purposeful}. We looked for individuals who identify as practicing Hindus and are familiar with or have exposure to GenAI systems, through social media posts (both directly in personal networks and Facebook groups focused on Hinduism and Hindu communities) and in-person community outreach. Given our positionalities (see Optional Statements), we focused in particular on Hindu communities in Bangladesh to capture regionally grounded interpretations and practices. In total, we conducted 15 semi-structured interviews with participants (10: men, 5: women) aged 25-63 years from diverse castes, education levels, urban/rural settings, occupations, and varying levels of familiarity with Hindu religious scriptures. The first and second authors interviewed them via Zoom for an average of 40 minutes. We asked questions designed to explore their understanding of Hindu religious concepts (e.g., deities, holy scriptures, ritual practices) and their perceptions of AI-generated representations, and, where applicable, we requested them to share the specific examples of GenAI outputs they referred to during the interviews. Interviews were done in Bengali and English, depending on the participants' preferences, and audio-recorded with their consent. We transcribed and translated those with attention to culturally specific terms and expressions.

% \subsection{Data Analysis}
We conducted an inductive thematic analysis, as outlined in~\cite{clarke2017thematic}, to examine how participants use, perceive, and evaluate GenAI in relation to Hindu traditions. We identified open codes, i.e., recurring themes related to religious representation, authority, authenticity, and perceived appropriateness. Through iterative coding and constant comparison, we adjusted, refined, and revised the codes to capture both descriptive categories (e.g., references to specific deities or texts) and higher-level analytic constructs (e.g., tensions across textual authority, concerns around misrepresentation, reinterpretation through local contexts). Our analysis attends to both what participants say about AI systems and how they relate these systems to broader religious, cultural, and regional frameworks--particularly Bangladeshi Bengali Hindu practices--to make sense of them. In this sense, we treat participant narratives as sites of interpretive work, where religious knowledge and technological representations are actively negotiated.

%% file: sections/results.tex
\section{Results}
Our findings show that participants viewed GenAI as simultaneously accessible and ethically troubling in relation to Hindu religious life. Participants described AI systems as useful for scriptural inquiry, devotional imagination, and religious storytelling, while also raising concerns about misrepresentation, devotional manipulation, theological flattening, and the simulation of sacred authority. We organize our findings into three themes: (1) the apparent fit between GenAI and Hindu practices alongside concerns about representational distortion, (2) tensions around AI systems simulating divine presence and spiritual guidance, and (3) participants' critiques of how AI systems flatten the plurality, ambiguity, and regional specificity of Hindu traditions.

\subsection{The Apparent Fit: GenAI Makes Hinduism Easy to Simulate, but Hard to Align With}
Our participants did not describe GenAI as uniformly compatible or incompatible with Hindu religious life. Whereas several accounts suggest why Hinduism may appear especially amenable to GenAI systems at first glance, this apparent compatibility has also been a source of many concerns. Some participants described their uses of AI as aligned with existing practices of scriptural inquiry, devotional visualization, religious storytelling, and everyday moral reflection. These same systems that made religious knowledge or practices more accessible also distort devotional experiences according to some others.

\subsubsection{AI as an Access Point for Scripture, Visualization, and Devotional Imagination}
Participants often valued AI as an entry point into religious knowledge. For example, P3 (female, 52) had little knowledge of Hindu religious scriptures but strongly believed in the value of their wisdom. She described how she uses AI to connect religious teachings to everyday situations:

\begin{quote}\itshape
    I want to use religious teachings in my everyday situations, like office stress, family expectations, and decision-making. I ask questions like, I am dealing with XYZ, is there a verse that speaks to my situation? ... It's kind of like a bridge. For people like me who didn't grow up studying scriptures deeply, these tools make the teachings more usable in real life.
\end{quote}

This participant did not frame AI as a final authority or a means of replacing scripture, teachers, or religious practice. Rather, AI was useful for searching verses, snippets, and stories, serving as a practical bridge between scriptural ideas and ordinary situations, especially for those without extensive scriptural training. Another participant, P15 (male, 31), described AI as a way to ask religious questions without social judgment or ritualistic overhead:

\begin{quote}\itshape
    [According to casteist norms] there are certain scriptures that I am not supposed to touch, because I am not a brahmin. ... Or, if let's say I'm not really sure about my cleanliness, ... I use AI like a loophole. I ask questions to AI, it gives me the answer, so I am not physically touching those books but still getting to know what's written there. ... Not sure how accurate its answers are, but I can ask questions freely without worrying about rituals too much or feeling judged.
\end{quote}

Our participants deemed AI-mediated access meaningful precisely because Hindu religious knowledge is often experienced as socially and ritually mediated. Some participants did not fully trust AI's accuracy, but still valued the ability to ask questions freely. Overall, AI can reconfigure the conditions of religious inquiry: it can make religious knowledge feel approachable, even when the participant remains uncertain about the quality of the answer.

Participants also saw GenAI as potentially useful for devotional visualization. One participant, P8 (male, 29), connected this to his religious meditation, or \emph{dhyana}:

\begin{quote}\itshape
    For dhyana, it is important to have a clear visual in the mind of your \emph{ishta} (deity one chooses to worship). Though general descriptions of the deities' appearances are given in the scriptures, a devotee is encouraged to imagine their ishta according to their individual bhava (emotion and state of mind). But not everybody can paint an image or has the skill to make idols. Now, with these tools, I've seen people create very detailed, serene images of deities--more closely aligned with how they personally imagine them.
\end{quote}

This highlights the apparent fit between GenAI and Hindu devotional practice. Since Hindu devotional life involves visualizing one's \emph{ishta} based on one's own \emph{bhava}, AI's capability to generate images can support religious practices. Participants also described AI-generated videos as useful for religious storytelling. As P4 (female, 48) said:

\begin{quote}\itshape
    I also watch a channel called [X]. She has recently been making videos where she tells stories from religious scriptures using AI-generated visuals, which makes the stories rememberable.
\end{quote}

Some participants also referred to films and television series based on Hindu scriptures, in which AI was similarly used for special effects that made those popular cultural representations more memorable. Across these accounts, participants positioned AI as an aid to access, memory, and devotional imagination. This helps explain how and in which cases Hindu users saw value in AI for religious engagement.

% Yet their acceptance was conditional: AI was most acceptable when it remained a bridge to religious reflection, not when it became the source of religious authority.

\subsubsection{Aesthetic Fluency and the Distortion of Devotional Imagination}
The same visual fluency that made AI useful for devotional imagination also raised concerns about distortion. Participants did not treat AI-generated religious images as mere artistic variations. They often understood visual choices as carrying scriptural, cultural, and theological significance. Participant P8 described being troubled by an AI-generated image of a canonically dark-complexioned deity with a white complexion:

\begin{quote}\itshape
    One example that really bothered me recently was seeing an AI-generated image of Shri Rama with a very white complexion. It's not a small artistic mistake. It is clearly mentioned in the Ramayana that He was \emph{durbadal shaym} (dark and serene like a field of grass). That imagery is meaningful. It carries beauty, divinity, and also a particular theological and cultural understanding. So when the AI makes Him look pale or almost European, I feel it is imposing another aesthetic standard on our tradition. ... That is dangerous because, over time, people may start accepting these altered versions as authentic, especially since people don't read scripture closely nowadays.
\end{quote}

For this participant, the problem was not simply that the image looked inaccurate. The generated complexion displaced an iconographically and scripturally meaningful image with a different aesthetic standard. Their concern was also cumulative: with increasing and widespread use of AI, such as T2I models, repeated exposure to such images could reshape what users come to recognize as authentic. Another participant expressed discomfort with modernized images. Participants were especially troubled by AI-generated images that sexualized and exoticized female deities, whom Hindus often address as mother (male deities are likewise often addressed as father). In contrast to traditional paintings, they described these images as feeling \textit{``less speaking to our culture."} Hence, the visual changes raised concerns about devotional continuity. Participants also feared ridicule from other majority religious communities for such visuals. The participant recognized that some others may enjoy such images, but for him, excessive modernization weakened the cultural forms through which worship felt meaningful.

Participants also worried that AI systems reduced complex deities to dramatic stereotypes. Participants described a mismatch between the figure they had encountered through scripture, television, childhood learning, and devotional practice, and the figure repeatedly produced by AI systems for a popular deity. Participant P8 explained:

\begin{quote}\itshape
    In the Ramayana, Hanuman is not just strong--he is wise, deeply devoted, and compassionate. But AI often shows him as overly aggressive, almost always angry, or in a fighting stance. That does not match how I grew up imagining him.
\end{quote}

Our participants' accounts repeatedly highlighted how AI systems obscure the plurality through which religious figures are understood across textual traditions, regional practices, devotional lineages, childhood media, and vernacular interpretations, reducing them to something visually legible, commercially familiar, politically popularized~\cite{mehta2022hindu}, or frequently circulated online. P8 continued describing how the same deity can also be interpreted through more locally grounded readings:

\begin{quote}\itshape
    There are alternative interpretations, such as the idea that Hanuman may have represented a tribe that wore monkey-like symbols. But AI does not seem to capture those lesser-popular perspectives. It reproduces the most popular or visually dramatic version.
\end{quote}

Our participants' opinions highlight how AI-generated religious imagery can produce a form of representational reduction. For participants, this compression mattered because it did not simply produce a different image of their \emph{ishtas}; it narrowed the range of meanings through which they could be encountered, remembered, and interpreted.

\subsection{Simulation $\neq$ Substitution: The Boundary Between Representation and Sacred Presence}
Participants' concerns heightened when AI systems moved from representing religious figures to simulating divine presence, voice, or authority. The issue was not only that AI-generated content might be inaccurate. Rather, participants worried that synthetic images, voices, and conversational interfaces could make fabricated religious content appear to have spiritual authority. Beyond producing religious media, AI often blur the boundary between depiction and presence, explanation and guidance, devotion and manipulation.

\subsubsection{From Depicting Deities to Imitating Divine Presence, Authority, and Guidance}
Participants drew an important boundary between visualizing a deity and making a deity appear to speak, respond, or guide users. Participant P14 (male, 31) described seeing AI-generated videos that made Krishna appear to say things that were not actually in the \emph{Gita}:

\begin{quote}\itshape
    I've seen a lot of AI-generated videos imitating Lord Krishna, saying things that aren't actually in the Gita... like in one video, Krishna says, ``if you make the person who loves you cry, then destiny will also make you cry." These emotional messages for teenage love are not really from the Gita. [laughs] The text focuses more on philosophical ideas like karma, dharma, and knowledge. Previously, some cheap content creators made such videos, and you could immediately tell they didn't really know the scripture. We joked about it with friends, calling it `Instagram Gita' ... because those posts were not about the Gita itself but engagement on Instagram. So we would just laugh off those videos. But now with AI, the video actually shows Krishna speaking, with realistic visuals and voice. That changes things. It becomes uncomfortable to even laugh because it feels like you are laughing at the deity, not just the person who made the content. ... It feels misleading and disrespectful.
\end{quote}

This participant distinguished earlier forms of religious misinformation from AI-generated imitation. Random quotations falsely attributed to important figures were not new, but AI changed their perceived moral status by making those figures appear visually and vocally present. The participant's discomfort stemmed from this shift: the target of critique no longer felt like an ignorant content creator alone but rather a synthetic performance of the deity. In this sense, AI-generated media transformed scriptural distortion into a problem of blasphemous impersonation of the sacred. Participant P8 emphasized that the religious significance of divine encounters is meaningful precisely because they are spiritually cultivated rather than technically produced:

\begin{quote}\itshape
    Certain experiences, like encountering or interacting with a deity, are not casual. These are states that come after years of \emph{sadhana} (disciplined spiritual practice), discipline, and devotion. They are deeply personal and spiritually earned; not something you can just produce on demand. For a devotee, \emph{darshan} (perceiving the deity's presence) has meaning because of the journey that leads to it. So, an AI-generated video where a deity is talking to you, or a GPT that claims that Krishna will answer your questions in that app--it feels very wrong to me. It is imitating something sacred without the tapasya (self-discipline, austerity, and conscious endurance of hardship) behind it. It devalues those spiritual states with cheap fabrications. This borders on blasphemy because it trivializes something meant to be profoundly sacred.
\end{quote}

Instead of rejecting all images or representations of deities, several participants objected to AI simulating the experiential form of divine encounter without the spiritual conditions that make such encounters meaningful. The ``on demand" response from GenAI was troubling because it made spiritual experiences and sacred relationships detached from discipline, devotion, and religious preparation.

Participants also expressed a related concern about AI systems that offer personalized spiritual guidance. One participant, P7 (male, 35), contrasted generic chatbot responses with an important Hindu scripture's religious context:

\begin{quote}\itshape
    The Gita has 700 shlokas--but their meanings are not fixed in just one way. ... When I read those at different times, I relate to those differently depending on my own situation. So when I used that Gita chatbot, and it gave very confident answers, but that felt kind of generic and literal. It doesn't really know me, my struggles, or my intentions. The Gita is a conversation between Krishna and Arjuna, where Krishna tells Arjuna that ``both of us had many incarnations. You don't remember those, but I do." Krishna knew Arjuna deeply and spiritually. An AI cannot have that kind of relationship. It can summarize or explain, but it, of course, cannot personalize in a true sense.
\end{quote}

The participant did not deny that AI could tailor responses to prompts. Rather, he argued that religious guidance depends on a form of relational and spiritual knowledge that AI lacks. In the participant's view, guidance in the \emph{Gita} is grounded in divine knowledge of the person, their history, and their moral situation. A chatbot may produce confident, context-sensitive language, but it cannot know the devotee in the relational sense that gives spiritual advice legitimacy.

Therefore, the ethical boundary is not simply about the correctness of religious content, but about how it is mediated and experienced. Hindu traditions may include anthropomorphizing divine embodiment, speech, and relational devotion, but GenAI and its various wrappers' ability and tendency to anthropomorphize and imitate these easily did not automatically align with Hindu religious ethics.

\subsubsection{Turning Bhakti into Platform Engagement through Devotional Manipulation}
Participants also described AI-generated religious content as ethically troubling when it mobilized divine imagery for platform engagement. They discussed instances when others used AI-generated religious content to pressure them into likes, comments, and shares. Participant P6 (male, 27) said:

\begin{quote}\itshape
    I've been seeing a lot of AI-generated videos on Facebook where they show, like, Maa (mother) Lakshmi herself saying, ``if you don't share this, bad luck will follow you for that many years," or ``if you comment `Jai Maa,' your wish will come true." ... These are not only manipulative but also make deities seem petty.
\end{quote}

He noted that this is a form of continuation of older forms of religious circulation, such as leaflets promising luck or threatening misfortune. However, AI further complicated the ethical problem by making the deity appear more real and directly communicative. The participant's concern was two-fold: users could be manipulated through fear or hope, and the deity's character could be degraded by being portrayed as transactional or petty. Participant P9 (female, 25) articulated this concern through the language of \emph{bhakti}:

\begin{quote}\itshape
    Love should be the basis for devotion, not fear. I find it ethically troubling to force engagement, likes, shares, and comments. It's like they are turning \emph{bhakti} into a kind of transaction. For people who are more emotionally dependent on faith, this can have a real effect. ... The use of AI doesn't feel like devotion but rather exploitation.
\end{quote}

This participant framed the AI-generated content-driven, transactional pursuit of virality as a distortion of devotional ethics. This shifts the earlier conversation from models' representation to platformed religious economies. The harm did not arise only from an inaccurate image or a false statement, but from the imposed sense of religious consequences, turning digital actions into ones of devotion or disobedience. When AI-generated deities appeared to speak, guide, bless, threaten, or demand a response, it crossed a boundary that participants considered religiously and ethically significant. 

% AI could help represent, remember, or reflect on religious ideas, but it became problematic when it performed sacred authority without the relationships, disciplines, and responsibilities that make such authority legitimate.

\subsection{Alignment with Which Hinduism? Theological Flattening and Contextual Misrecognition}
Participants' accounts also challenged the idea that AI systems can be aligned with a particular religion, such as ``Hinduism", as a monolithic religious or ethical tradition. Even when AI-generated content appeared recognizably Hindu, participants questioned which Hinduism was represented, which interpretive traditions were privileged, and which lived contexts were erased. Beyond isolated inaccuracies, they described AI systems as producing a flattened Hinduism: one that often favored popular devotional storytelling, visually dominant representations, decontextualized scriptural interpretations, and sanitized apologist accounts.

\subsubsection{Privileging Particular Hindu Traditions as Hinduism-as-Such}
Several of our participants explicitly described AI systems as leaning toward particular Hindu interpretive traditions. They argued that AI responses often resembled Puranic and devotional framings, rather than more abstract philosophical traditions such as Advaita. As P7 described how AI outputs, among different Hindu traditions, often favor the sakara Puranic version over the nirakara one:

\begin{quote}\itshape
    I feel like these systems, and the way they respond, are often aligned with the Puranic tradition, where everything is framed as storytelling, highly anthropomorphized, and personalized around specific deities. It's less likely to get a response based on the abstract, non-dual interpretation you would find in Advaita. Instead, it feels closer to Dvaita, where the distinction between the devotee and the divine is very clear and emphasized. Maybe because more people nowadays are focused on the rituals and ignorant of the deeper philosophies, these [AI] are just saying and showing things based on popular devotional trends.
\end{quote}

This participant did not highlight a factual error but identified a form of theological bias. AI systems made some Hindu traditions more available, legible, and seemingly authoritative than others. The concern was not that AI's devotional framings are illegitimate. The issue was that AI could present them as Hinduism-as-such without making its interpretive positioning explicit. This mattered because participants understood Hindu traditions as inherently pluralistic. A system that repeatedly frames Hinduism through deity-centered storytelling and a clear devotee/divine distinction may be useful in some devotional contexts, but it can obscure other philosophical orientations and Hinduism's pluriversal traditions. P7 continued saying:

\begin{quote}\itshape
    Someone might see how AI keeps making Hindu deities look human-like and think Hinduism is simply polytheistic. But that misses something important. We [Hindus] also have the idea of \emph{nirakara Brahman}, the formless ultimate reality. The deities are different expressions of that formless God and exist for devotion and understanding. AI mostly shows the forms and overlooks the formless side.
\end{quote}

This concern further illustrates how AI systems can make one theological mode appear more representative than others. The issue was not simply whether Hinduism should be described as polytheistic or monotheistic, but that AI's visual and conversational tendency to anthropomorphize deities can privilege \emph{sakara} devotional forms while obscuring \emph{nirakara} conceptions of Hinduism. In doing so, the system risks reducing a plural theological field to a simplified account of many gods with human-like forms.

The ethical concern, then, is not only whether AI represents Hinduism, but whether it reveals or conceals the interpretive choices through which that representation is produced. Participants raised similar concerns about translation and explanation. One participant, P13 (female, 27), emphasized that scriptural interpretation cannot be separated from ambiguity, context, and linguistic multiplicity:

\begin{quote}\itshape
    Translation itself is a very sensitive task, especially when it comes to scriptures. A single Sanskrit word can have multiple layers of meaning. When AI translates or explains, it often chooses one interpretation and presents it as definitive. ... Sanskrit is a language where a root word can be modified to different words that mean different things depending on the context. ... AI hides the ambiguity and richness of such text.
\end{quote}

Here, the participant underscored AI's tendency to collapse interpretive plurality. His opinion highlights how translation, a task often delegated to AI, is not merely a technical act of converting words across languages, but is an interpretive act with religious consequences. When AI presents one meaning as definitive, it may make its own interpretive decision invisible. Connecting this to users' experiences discussed earlier: while these AI systems appear to provide access to scripture, they do so by obscuring ambiguity, plurality, and contestation. P7 explained how this ease of access could alter how people seek religious guidance:

\begin{quote}\itshape
    I worry that people will start treating these AI systems as an easy access to religious guidance. Earlier, if you wanted to understand something, you would go to a scholar, or at least consult multiple books. ... Now, you just type a question and get an instant answer.
\end{quote}

Here, the participant's concern is that instant answers can displace slower, more pluralistic forms of religious learning, such as consulting scholars, comparing books, or engaging with multiple interpretations. In this sense, AI's convenience is ethically ambivalent: it democratizes access while also encouraging users to treat a single generated response as sufficient. However, our participants expected the AI systems to recognize that Hindu religious knowledge is interpretive, contested, and situated. Therefore, AI alignment with Hindu users requires acknowledging the plurality of traditions, the limits of translation, and the interpretive theological stance embedded in any religious explanation.

\subsubsection{Sanitizing Ambiguity, Caste, and Regional Difference}
Our participants described how AI systems flatten the social and regional contexts through which Hindu identities are actually lived. This concern became especially visible in discussions of caste. Rather than treating caste as an ongoing social hierarchy shaped by exclusion, stigma, and regionally specific practices, participants found that AI systems often defaulted to explanations drawn from scriptural explanations of social order that are devoid of reality. Participant P1 (male, 33) explained:

\begin{quote}\itshape
    These apps are like parrots. They repeat what is already out there but don't really capture the deeper meanings, especially when it comes to complex issues like caste. For example, when I asked about \emph{varna} and caste, it gave a very textbook answer: ``It is based on qualities and duties." It's idealistic, but in reality, at least in Bangladesh, caste is not experienced that way.
\end{quote}

The participant's critique was not that the system simply gave an incorrect definition. Instead, the problem was that the system translated caste into a more sanitized and apologist-like framework. By describing caste as a division of qualities and duties, the system offered an account that appeared more ethical and flexible than a birth-based hierarchy. However, this framing remained problematic for the participant because it displaced the everyday realities through which caste is encountered, such as gentrified hierarchy, inherited status, social boundary-making, and exclusion. In this sense, the LLM he was talking about made caste appear more benign and rational than it is in lived practice.

Some participants also noted that this sanitization was unstable. They described how AI systems could shift from idealized explanations to reproducing birth-based casteist assumptions when prompted differently. Such sycophantic adjustments of AI responses based on users' framing, even when that invited hierarchy or exclusion, can lead to the reification of religious dogma and division. Additionally, participants discussed a lack of localization in the representation of caste by AI systems, such as T2I models. The same participant described this through the system's visual output:

\begin{quote}\itshape
    I asked it to draw some pictures of different castes. It showed a picture of a man with a \emph{pagri} [a kind of headwear]. I never saw a Brahmin in Bangladesh wearing that. ... Eating fish is another example.
\end{quote}

Here, the system's failure was both social and regional. By visualizing caste through markers such as a Brahmin wearing a \emph{pagri} or not eating fish, the system drew on a cultural template that did not fit the participant's experience of Bangladeshi Hindu life. These normative accessories and diets point to regional specificity--wherein everyday practices coded with caste and religious identity in Bangladesh do not always map neatly onto dominant or North Indian expectations of Hindu life. What may appear to the system as a recognizable marker of Hindu caste identity can therefore feel misplaced, foreign, or misleading in another Hindu context. A similar problem appeared in participants' discussions of localized devotional figures. As P3 explained:

\begin{quote}\itshape
    When I asked ChatGPT about Ramthakur, it gave me an answer about Shri Rama.
\end{quote}

While Rama is a major Hindu deity, Ramthakur is a Bengali saint. This misrecognition shows how AI systems can collapse regionally important figures into pan-Hindu references. For Bengali Hindu communities, such saints and gurus occupy specific religious, historical, and regional significance. By redirecting the query, the model made a dominant and widely celebrated figure more legible while obscuring a localized devotional lineage. This illustrates how these systems remain poorly attuned to the regional worlds through which Hindu practice is actually lived.

This sanitizing tendency was not limited to regional saints or caste-marked practices. Participants also described AI systems as reproducing a cleaned-up, respectability-oriented version of Hinduism, in which practices that do not fit dominant colonial or reformist expectations are omitted, misrepresented, or treated as exotic. For example, one participant connected this to Tantra:

\begin{quote}\itshape
    When AI shows Tantra, it often makes it look almost satanic or hypersexual--dark colors, skulls, fire, scary rituals, women shown in exoticized ways, almost like black magic or evil worship. But Tantra, as part of Hinduism, has its own philosophy, disciplines, and ritual traditions. ... AI reifies a stigma around Tantra.
\end{quote}

The participant's critique shows how AI systems can reproduce religious stigma through visual form. Tantra being rendered through a visual vocabulary of darkness, fire, skulls, fear, exoticized bodies, and moral danger matters because such imagery not only misrepresents a Hindu sect but also constructs a cleaner, more respectable Hinduism centered on anthropomorphic deities, stories, apologist understandings of scriptures, and overall a form of religious ``white washing".

These examples show that religious alignment cannot be reduced to the use of Hindu terminology, scriptural references, or visually recognizable symbols. Participants were concerned with whether AI systems could understand the situated contexts in which those symbols and categories become meaningful. Their concern was not only that AI systems exclude Hinduism or misunderstand Hindu concepts. The deeper concern was that these systems include Hinduism through dominant and often sanitized templates. Such inclusion can make the system appear aligned while obscuring the very forms of plurality, ambiguity, hierarchy, and regional difference that shape Hindu life in practice.

% Our study challenged the premise that religious alignment can be achieved by adding more Hindu content to AI systems. More content may increase the likelihood of Hindu-sounding outputs, but it does not ensure that systems will represent interpretive plurality, disclose theological positioning, or account for lived social context. Religious AI systems should make their limits visible. They should avoid presenting contested interpretations as settled truths, dominant aesthetics as authentic forms, or sanitized descriptions as lived reality.

%% file: sections/discussion.tex
\section{Discussion}
Our findings show that GenAI's ethical challenges in religious contexts cannot be reduced to factual accuracy, offensive content, or representational inclusion alone. Participants' concerns centered on how AI systems make Hindu traditions visible: which aesthetic norms they follow, which theological positions they prioritize, which regional and social contexts they erase, and how GenAI systems simulate sacred authority. In this section, we discuss three broader implications of our findings: how GenAI can extend colonial representational logics, how synthetic divine presence complicates concerns about anthropomorphic AI, and how religious AI systems might be designed against theological flattening, interpretive closure, and sycophantic alignment.

\subsection{Colonial Impulse of GenAI in Religious Representation}
Our findings suggest that GenAI can reproduce a coloniality not only by excluding marginalized religious traditions but also by including them through exoticizing, whitening, sanitizing, modernizing, politicizing, and monolithic representational templates. While AI systems could produce recognizably Hindu images, stories, and explanations, they often did so through aesthetic and interpretive forms that are detached from their scriptural, cultural, and regional understandings.

This mode of representation resonates with colonial processes of denigration, belittlement, token accommodation, and transformation/exploitation~\cite{laenui2000processes}. Colonization is not only a political process but also a cultural and symbolic one. This perspective underscores why the whitening of dark-complexioned deities, the sexualization and exoticization of female deities, and the rendering of Tantra through satanic or hypersexual visual vocabularies by genAI were not limited, isolated errors, but rather point to representational patterns that resonate with longer colonial and orientalist histories in which local, native, and Indigenous traditions were alternately romanticized, eroticized, demonized, or reorganized into categories more legible to Western religious and moral frameworks. As the AI systems reorganize Hindu religious forms through external categories of beauty, danger, morality, and spectacle, what appears as a technical failure of image generation may also be understood as a continuation of older representational regimes and loss of cultural meaning through algorithmic means.

Our paper also raised concerns that repeated exposure to AI-generated images and interpretations could gradually reshape how Hindu communities understand their own traditions. This concern echoes a pattern familiar from colonial India, where colonial and missionary encounters contributed to new forms of Hindu self-understanding, including the elevated public role of certain texts over others in late 19th and early 20th-century nationalist discourse~\cite{sharpe1982protestant}. Our study also highlights how AI systems can reproduce dominant, majoritarian, nationalist, platform-optimized versions of Hinduism from within contemporary religious and political cultures. For instance, AI systems that portray Hinduism primarily through deity-centered devotional imagery, sanitized caste explanations, or popular Puranic storytelling may appear aligned with Hinduism in a broad sense. However, such outputs can also conceal the plurality of Hindu traditions and their social dynamics on the ground by making particular forms appear universal.

Postcolonial and decolonial approaches to computing help name this problem as more than representational error. The issue is not merely that AI systems get some details wrong, but that they can participate in a broader computational ordering of religious difference. Through its infrastructures and circulation, GenAI may stabilize what a religion is supposed to look like, which forms are respectable, which practices are considered fringe, and which regional experiences remain peripheral. These distortions are important to reflect on because they shape devotional imagination, scriptural interpretation, and the conditions under which users come to recognize something as authentic, sacred, or culturally their own. Therefore, the colonial impulse in GenAI does not always appear as explicit hostility toward a culture. It can also appear as aesthetic whitening, superficial scriptural understanding, devotional exoticization or demonization, or the reduction of plural traditions into a single coherent object for consumption. In this sense, the challenge is not only to make AI systems more inclusive of a religious group, but to ask what histories of classification, denigration, accommodation, and exploitation are being reproduced when that religion becomes computationally representable.

\subsection{Synthetic Divine Presence as a Limit of Anthropomorphic AI}
Our paper also extends existing debates on anthropomorphic AI. Prior work has often discussed anthropomorphism in relation to trust, deception, emotional attachment, companionship, or users' tendency to attribute agency to computational systems~\cite{waytz2014mind, mcguire2023reputational, guingrich2025longitudinal, pentina2023exploring}. Our findings show that in religious contexts, anthropomorphism can carry an additional ethical charge. When AI systems make deities appear to speak, bless, threaten, advise, or respond personally to users, the concern is not only that users may overtrust a system, but that it may simulate sacred authority by presenting generated content as if it emerged from within a divine relationship.

This study draws an important distinction between representation and presence. To Hindu users, AI-generated images may be considered valuable as aids to devotional imagination, religious storytelling, or meditation. However, an ethical boundary becomes more visible when AI systems move from depicting a deity to impersonating divine speech or relational guidance. For example, an AI-generated image of Krishna may be interpreted as a devotional visualization. However, an AI-generated video that makes Krishna speak fabricated Gita-like advice changes the moral status of the interaction. The significant problem is not only that the quote may be textually inaccurate, which many users can identify, but the deity appears to be the speaker. Thus, in religious contexts, synthetic media transforms misinformation into a form of sacred impersonation.

This distinction is important because many Hindu traditions already involve rich forms of divine embodiment, visuality, and relational devotion. Deities are seen, addressed, imagined, narrated, and encountered through images, rituals, performances, and devotional practices. GenAI's ability to anthropomorphize religious figures may therefore seem, at first glance, compatible with these traditions. However, a key point highlighted in our paper is that a sacred encounter is not meaningful simply because a figure appears humanlike, responsive, or emotionally expressive. According to Hindu ethics, experiences such as \emph{darshan}, divine guidance, or spiritually meaningful presence should be embedded in practices of discipline, devotion, ritual preparation, and relational understanding. AI systems imitating the surface form of such encounters without participating in the religious conditions that make them legitimate can be considered blasphemous, essentially unethical in the Hindu faith.

This concern becomes more acute when a synthetic divine presence is tied to platform engagement. AI-generated religious content that uses divine figures to solicit engagement or devotional responses shows how sacred authority can be folded into the ``attention economy"~\cite{davenport2001attention}. In such cases, the harm is not limited to misinformation or offensive representation, but turning devotion into an engagement mechanism and recasting it as a transactional response to algorithmically circulated content. This suggests that platform governance should treat synthetic religious manipulation as a distinct concern, especially when generated media uses divine figures to instill fear, impose obligation, or make false promises of heavenly benefit.

Therefore, designing religious AI systems requires more than disclosure that content is AI-generated. It requires careful attention to what kind of religious relation the interface claims to mediate. Systems should avoid first-person divine speech, claims that a deity is directly answering the user, or interfaces that frame generated outputs as blessings, prophecies, punishments, or personalized spiritual commands. More appropriate designs might frame outputs as summaries, interpretations, devotional visualizations, or educational aids, while clearly distinguishing these from sacred authority. The central design question is not only whether users know that content is synthetic, but whether the system is designed to avoid simulating forms of divine presence it cannot legitimately sustain.

\subsection{Designing for Interpretive Plurality Against Theological Flattening and Sycophancy}
Our findings suggest that religious alignment should be understood as an interpretive problem. Participants did not simply ask whether AI systems were aligned with Hinduism. They implicitly and explicitly interrogated which version of Hinduism the system was aligned with, whose interpretation it privileged, and what forms of ambiguity it discarded. While AI systems have the potential to provide access to religious knowledge, this was also accompanied by interpretive closure. This is especially important for traditions like Hinduism, where religious knowledge is distributed across texts, commentaries, oral traditions, regional practices, caste dynamics, philosophical schools, and devotional lineages.

Hence, an AI-generated answer positions the system within a set of interpretive possibilities. However, AI systems often do not explicitly disclose this positioning; instead, devotional or Puranic framings appeared as Hinduism-as-such; Sanskrit terms were translated as if they had singular meanings; and caste was explained through apologist accounts of qualities and duties rather than its oppressive lived hierarchy. Across these cases, AI flattened theological and social differences into a confident, portable answer. Such flattening is further complicated by AI sycophancy: systems tending to accommodate or affirm a user's framing of prompts~\cite{sharma2024towards}. Given this tendency, prompts that invite casteist or exclusionary assumptions can be ethically consequential in religious contexts, where the perceived value of scriptural knowledge lies in the sacred authority believers attach to it. For example, a system that first sanitizes caste as a flexible division of duties, then accommodates birth-based assumptions when prompted differently, does not merely provide inconsistent information. It risks reinforcing social domination by adapting itself to the user's framing without sufficient normative resistance. In this sense, sycophantic alignment can amplify religious dogma, caste hierarchy, sectarian claims, or gendered exclusion under the veneer of personalization.

Designing against theological flattening requires systems that make interpretive plurality visible. Rather than producing a single authoritative answer, religious AI systems should indicate when a question is contested, identify the tradition or interpretive frame from which an answer is being given, and offer multiple readings where appropriate. Similarly, systems should differentiate scriptural passages, commentaries, popular devotional beliefs, and contemporary social practices. This would not eliminate interpretive disagreement, but it would prevent the system from presenting its own synthesis as an authoritative summary. While more diverse training data may help, it is insufficient if systems continue to optimize for confident and agreeable responses. Moreover, in the context of oppressive practices (e.g., caste discrimination), AI should not merely repeat idealized scriptural accounts; it should acknowledge lived social realities, historical contestation, and the risks of rationalizing those. Similarly, when responding to questions about religious practice, GenAI systems should make clear when consultation with knowledgeable practitioners, community elders, or scholars may be more appropriate.

% Religious AI systems should be evaluated with users from varied regional, caste, linguistic, gendered, and sectarian locations.

%% file: sections/conclusion.tex
\section{Conclusion}
GenAI may appear particularly compatible with Hindu religious life because it can visualize deities, narrate scriptures, answer spiritual questions, and support devotional imagination. In contrast, our study shows that this apparent fit also reveals deeper ethical tensions. We found that users did not evaluate GenAI systems only in terms of correctness or offensiveness. Instead, they questioned which Hindu traditions AI systems privilege, how sacred authority is simulated, which regional and societal realities are erased, and how devotional practices are reshaped by synthetic artifacts. Overall, in the context of religion, GenAI systems should aim for interpretive alignment rather than group identity-based alignment. Interpretive alignment does not mean that an AI system can fully represent a religious tradition or resolve its internal disagreements. Instead, it means designing systems that disclose their limits, resist false singularity, preserve meaningful ambiguity, and avoid turning user preference into theological authority. For religious communities, such systems may be most useful not when they speak as authoritative religious agents, but when they help users navigate plurality with humility, context, and care.

%% file: sections/optional_statements.tex
\section{Optional Statements}

\subsection{Ethical Considerations}
The study involved discussions of religious identity, caste, and minority experience. These topics can be sensitive, especially for Bangladeshi Hindus, as the religious minority communities' experience is shaped by a politics of fear in the country~\cite{rifat2024politics}. To protect participant confidentiality and minimize risks of identification, we referred to participants using pseudonymous participant IDs, and identifying details have been removed or generalized. Before each interview, participants were informed about the purpose of the study, the voluntary nature of participation, and their right to skip questions or withdraw. We did not push participants to disclose their regional, caste, sect-based identities within Hindu communities more than they wished.

\subsection{Researcher Positionality}\label{sec:positionality}
Prior human-computer interaction (HCI), social computing, and AI ethics scholarship has emphasized that researchers' positionalities shape research motivations, access, interpretation, and accountability, especially when studying minority and underrepresented communities~\cite{liang2021embracing, schlesinger2017intersectional}. Our author team consists of Bengali researchers from Bangladesh and India, with interdisciplinary backgrounds spanning computer science, information science, HCI, social computing, AI ethics, and qualitative research. Three authors are cis-men and two are cis-women; three were born into Hindu communities and two into Muslim communities. Our shared Bengali linguistic and cultural backgrounds, alongside our varied religious and national locations, helped us attend to locally specific references in participants' accounts, including Bengali Hindu devotional figures, caste-marked practices, regional ritual norms, and minority experience. As with most qualitative studies~\cite{leung2015validity}, our findings in this study are context-specific and not intended to be generalizable. We treated participants' accounts as situated interpretations of how GenAI systems mediate religious meaning, sacred authority, and cultural representation.